\documentclass[letterpaper, 10 pt, conference]{ieeeconf}  % Comment this line out if you need a4paper
\usepackage{graphicx}
\usepackage{amsmath}   % 基础数学环境（必须）
\usepackage{amssymb}   % 提供数学符号（推荐）

\IEEEoverridecommandlockouts                              % This command is only needed if 
\title{\LARGE \bf
High-Precision Climbing Robot Localization Using Planar Array UWB/GPS/IMU/Barometer Integration
}

\author{Shuning Zhang$^{1}$, \textit{Graduate Student Member, IEEE},
Zhanchen Zhu$^{2}$, Xiangyu Chen$^{1}$, \textit{Student Member, IEEE}, \\Yunheng Wang$^{1}$, Xu Jiang$^{1}$, Peibo Duan$^{3}$, Renjing Xu$^{1*}$
\thanks{*This work was supported by ...(corresponding author: Renjing Xu)}% <-this % stops a space
\thanks{$^{1}$Shuning Zhang, Renjing Xu, Xiangyu Chen, Yunheng Wang and Xu Jiang are with the Hong Kong University of Science and Technology (Guangzhou), Guangzhou 511458, China
        {\tt\small (szhang272@connect.hkust-gz.edu.cn, renjingxu@hkust-gz.edu.cn, xchen519@connect.hk
        ust-gz.edu.cn,  yunhengwang1214@gmail.com, xjiang159@connect.hkust-gz.edu.cn )}}%
\thanks{$^{2}$ Zhanchen Zhu is with the Department of Data Science and AI,
Monash University, Melbourne, VIC 3800, Australia  {\tt\small (Zhanchen.
Zhu@monash.edu)}}
\thanks{$^{3}$ Peibo Duan is with the Department of Data Science and Artificial Intelligence, Faculty of Infomation Technology, Monash University, Suzhou, Jiangsu, China  {\tt\small (peibo.duan@monash.edu)}}
}

\begin{document}

\maketitle
\thispagestyle{empty}
\pagestyle{empty}

%%%%%%%%%%%%%%%%%%%%%%%%%%%%%%%%%%%%%%%%%%%%%%%%%%%%%%%%%%%%%%%%%%%%%%%%%%%%%%%%
\begin{abstract}

To address the need for high-precision localization of climbing robots in complex high-altitude environments, this paper proposes a multi-sensor fusion system that overcomes the limitations of single-sensor approaches. Firstly, the localization scenarios and the problem model are analyzed. An integrated architecture of Attention Mechanism-based Fusion Algorithm (AMFA) incorporating planar array Ultra-Wideband (UWB), GPS, Inertial Measurement Unit (IMU), and barometer is designed to handle challenges such as GPS occlusion and UWB Non-Line-of-Sight (NLOS) problem. Then, End-to-end neural network inference models for UWB and barometer are developed, along with a multimodal attention mechanism for adaptive data fusion. An Unscented Kalman Filter (UKF) is applied to refine the trajectory, improving accuracy and robustness. Finally, real-world experiments show that the method achieves 0.48 m localization accuracy and lower MAX error of 1.50 m, outperforming baseline algorithms such as GPS/INS-EKF and demonstrating stronger robustness. 

\end{abstract}

%%%%%%%%%%%%%%%%%%%%%%%%%%%%%%%%%%%%%%%%%%%%%%%%%%%%%%%%%%%%%%%%%%%%%%%%%%%%%%%%
\section{INTRODUCTION}

High-altitude operations, such as inspection and maintenance of wind turbine blades [1-3], glass curtain wall cleaning [4-5], and rust removal of steel structures [6-8], have long been challenges in industrial applications. Due to their high-risk nature, numerous researchers and companies have conducted studies and research on intelligent robots for high-altitude work [9-11]. Common scenarios are depicted in Figure 1.

\begin{figure}[htbp]
    \centering
    \includegraphics[width=0.48\textwidth]{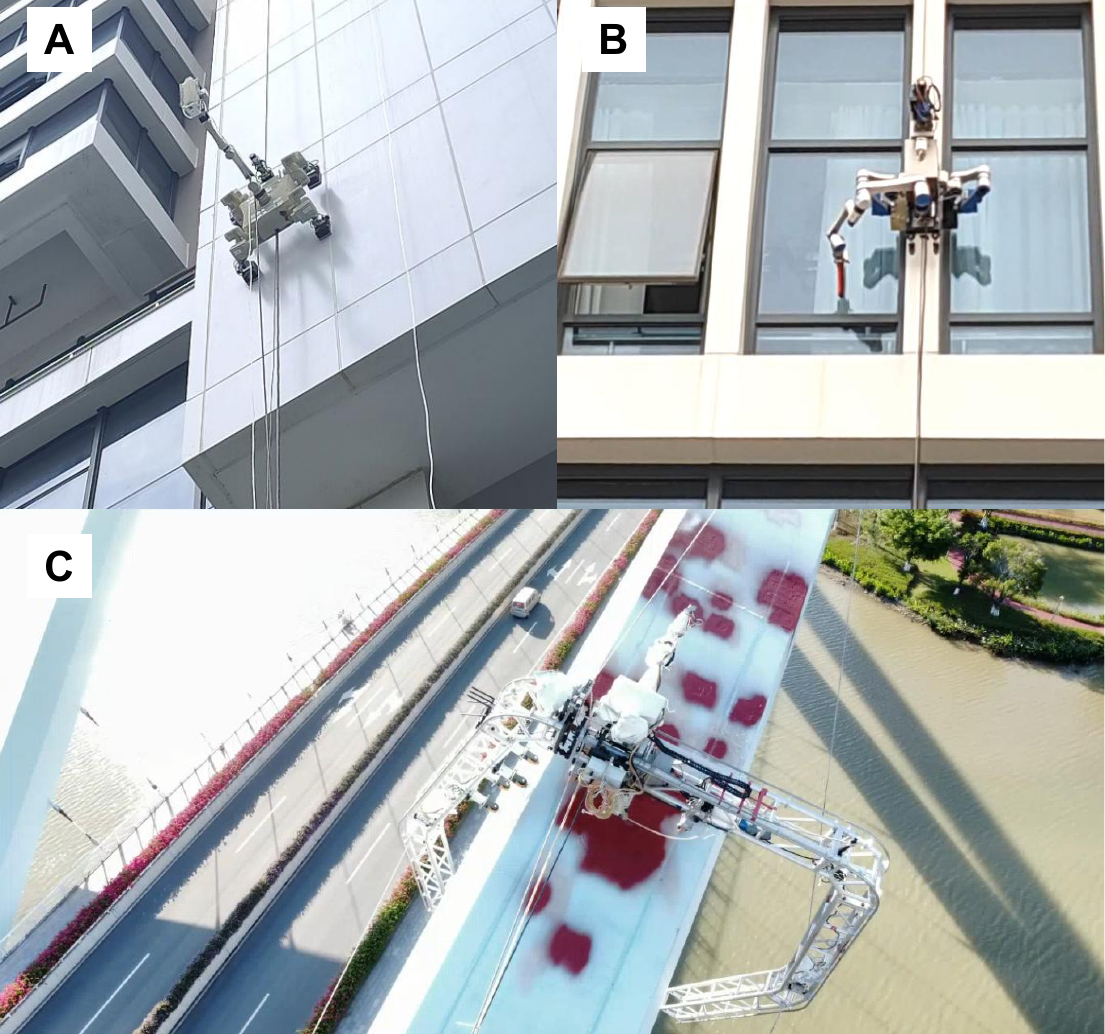}
    \caption{High-altitude operation robot scenarios. A presents building exterior wall cleaning. B illustrates high-altitude window cleaning. C depicts steel bridge rust removal and painting. }
    \label{f1}
\end{figure}

In high-altitude operation scenarios, centimeter-level localization capability directly determines the operational accuracy and stability of climbing robots [12]. It not only mitigates the risk of the equipment falling but also provides a reliable basis for path planning and collision avoidance in complex environments [13]. For precision tasks such as curtain wall installation, localization accuracy from the sub-meter to the centimeter level is a prerequisite for ensuring the quality of the work [12, 14]. Moreover, in multi-robot collaborative operations, high-precision localization is the foundation for achieving coordination [15]. In environments where GPS signals are limited, a localization solution that fuses UWB and IMU can effectively overcome the limitations of single sensors, providing the robot with a continuous and stable spatial reference [16-17] and significantly enhancing the reliability and adaptability of high-altitude operations.

However, achieving such high-precision localization faces significant challenges in typical climbing scenarios. Robots often operate on the vertical surfaces of large buildings or metal structures. In these partially occluded environments, GPS signals can be severely distorted due to multipath effects, failing to provide continuous and reliable localization information [17-18]. Inertial navigation System (INS) can provide high-frequency motion and attitude information, but its inherent integration drift causes errors to accumulate indefinitely over time [19-20]. The robustness and accuracy of visual localization solutions (like VIO or SLAM) are also difficult to guarantee when facing texture-less surfaces such as glass curtain walls or uniformly painted walls, often resulting in tracking failure [20]. Electromagnetic interference in environments with large steel structures (such as oil storage tanks, bridges, ship hulls) can significantly affect the accuracy of magnetometer information, reducing the reliability of localization or heading estimation methods that rely on the geomagnetic field [21]. Furthermore, climbing robots have strict requirements for payload and energy consumption, making it difficult to deploy some power-intensive sensors, such as high-performance 3D LiDAR.

Considering these challenges, researchers have explored various localization solutions. Vision-based methods, such as the classic VINS-Mono framework [22], are widely used in general robotics but show shortcomings in the unique environments faced by climbing robots [20]. To overcome the limitations of vision, researchers have investigated other sensors, for example, using rope length to provide global constraints [2] or using LiDAR to build point cloud maps [23]. However, these methods either restrict the robot's freedom of movement or are unsuitable for lightweight platforms due to issues with cost and power consumption. Therefore, multi-sensor fusion and lightweight design are the core future directions for climbing robot localization technology [24], and deep learning has also shown great potential for enhancing localization accuracy and system robustness in complex environments [25].

In summary, while many researchers have explored the positioning of high-altitude climbing robots using sensors like GPS, IMU, vision, and LiDAR, key challenges remain: low accuracy and robustness, poor environmental adaptability, and high energy consumption. Based on these issues, this paper models the localization scenario for high-altitude climbing robots and proposes a multi-sensor fusion architecture, aiming to enhance the accuracy and robustness of the climbing robot's localization system. The key contributions of this paper are as follows:

\begin{itemize}

\item A high-precision localization architecture based on the planar array UWB/GPS/IMU/Barometer integration was proposed, significantly improving accuracy and robustness in partially GPS-occluded environments. Additionally, this combination is currently novel in the field of high-altitude climbing robot localization.
\item Specifically for the multi-sensor configuration discussed in this study, a localization problem model was developed for high-altitude climbing robots.
\item End-to-end neural network models were proposed for both planar array UWB and barometer solution.
\item  An efficient fusion model, based on the attention mechanism, was proposed to achieve data-driven and adaptive multimodal fusion.

\end{itemize}

\section{Related Works}

In addition to the research on climbing robot localization mentioned in the I, related works also include UWB localization systems and multi-sensor fusion localization systems.

\subsection{Research on UWB Positioning Systems}

UWB technology shows centimeter-level potential in the field of localization due to its high-precision and strong-penetration ranging capabilities. However, its performance is highly dependent on line-of-sight (LOS) propagation, making NLOS propagation a core challenge [26]. In recent years, a popular direction has been to delve into its physical layer information and use machine learning methods to fundamentally address performance bottlenecks [27]. Furthermore, Through error analysis of UWB Time of Arrival (TOA) and Time Difference of Arrival (TDOA) models, the algorithm flow integrating UKF, Feedforward Neural Network (FNN), and Redundant Information Correction (RIC) achieves millimeter-level static accuracy and centimeter-level dynamic precision [28, 29].

UWB antenna array systems estimate the Angle of Arrival (AOA) by measuring the Phase Difference of Arrival (PDOA) of signals at different antenna elements, which supports the simultaneous measurement of both angle and distance [30]. However, traditional PDOA algorithms are susceptible to non-ideal hardware factors, limiting their measurement accuracy [31]. To address this issue, Ben Van Herbruggen et al. [32] proposed a CIR-based angle estimation method, combining AoA with Two-Way Ranging (TWR) to enhance system robustness. David Märzinger et al. [33] adopted a time-multiplexed AOA scheme to reduce hardware costs through antenna switching with a single receiver chain. Meanwhile, Nikita I. Petukhov's team [34] utilized an IMU-assisted Pedestrian Dead Reckoning (PDR) algorithm to solve the phase ambiguity problem in long-baseline systems.

\subsection{Research on Multi-sensor Fusion Architectures}

Multi-source data fusion aims to combine the advantages of different sensors to obtain information that is more accurate and reliable than what can be achieved with a single sensor [18, 35]. In recent years, fusion architectures have been evolving from classic probabilistic filtering models toward more intelligent and robust frameworks, such as factor graphs and machine learning.

Factor graph optimization methods achieve global trajectory optimization by maintaining states and measurements within a time window. Kenji Koide et al. [36] proposed a sliding-window tightly coupled method that combines point cloud registration with IMU data. This approach maintains robust pose tracking even under extreme conditions like point cloud degradation or sensor interruption. For high-precision inertial navigation, Pin Lyu et al. [37] improved the IMU pre-integration model by incorporating factors like the Earth's rotation and curvature. 

In terms of machine learning frameworks, the recently proposed Adaptive Kalman-Informed Transformer (A-KIT) framework [38] uses a Set-Transformer network to optimize the process noise covariance matrix in real-time. Additionally, the Transformer model and its core attention mechanism provide an ideal solution for multi-source information fusion [39-40]. The core idea of the cross-modal attention mechanism is to dynamically allocate information weights, enabling the model to achieve adaptive fusion. This concept of dynamic, adaptive weighting is well-suited to address the core challenge for climbing robots localization with multi-sensor systems.

\section{System Modeling and Problem Formulation}

The localization scenarios for high-altitude climbing robots differ significantly from those of ground robots. Their movement is primarily vertical with relatively little horizontal motion, which demands higher accuracy in vertical localization. Furthermore, these robots operate on the exterior surfaces of structures like buildings and bridges, creating a semi-occluded environment for GPS signals. This significantly impacts the accuracy and robustness of the GPS system. Therefore, it is essential to establish a dedicated localization model for climbing scenarios, as illustrated in Figure 2 of this paper.

\begin{figure}[htbp]
    \centering
    \includegraphics[width=0.48\textwidth]{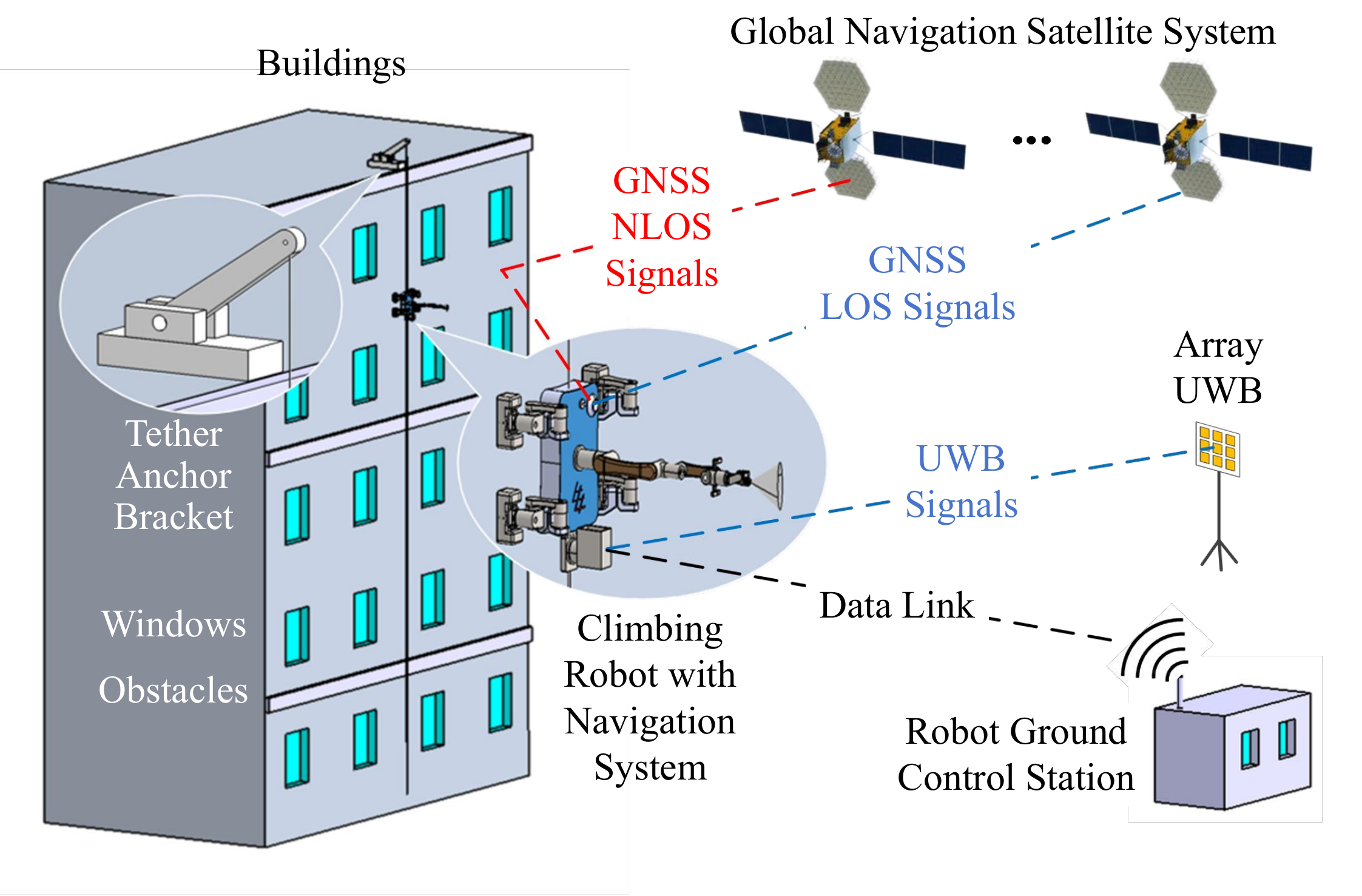}
    \caption{Localization scenario model for high-altitude climbing robot}
    \label{f2}
\end{figure}

Figure 2 illustrates the operational scenario of a high-altitude climbing robot working on the exterior wall of a building and its multi-sensor fusion localization system. The robot utilizes an adsorption-based locomotion mechanism for autonomous movement on vertical surfaces. It is equipped with a safety tether system and communicates with a ground station via a wireless link to ensure operational safety and remote monitoring. The localization system consists of a GPS receiver, an IMU, a planar array UWB  transmitter module, and a barometer, all mounted on the robot's body. This system, combined with a UWB receiving base station deployed in the environment, forms a hybrid localization architecture: GPS provides absolute localization, the IMU performs short-term dead reckoning, the planar array UWB achieves single-anchor localization through joint ranging and direction estimation, and the barometer compensates for vertical altitude drift.

Based on the scenario definition and the localization system configuration for the climbing robot, the precise localization problem for a high-altitude climbing robot is defined as follows:

\noindent \textbf{Problem 1}: In three-dimensional complex climbing scenarios, given the localization system configuration of a climbing robot (planar array UWB/GPS/IMU/barometer Integration), the observed state parameters are obtained as shown in Equation (1).

\begin{subequations}\label{eq:sensor_measurements}
\begin{align}
    \boldsymbol{Z}_{t}^{\mathrm{IMU}} &= \begin{bmatrix} \boldsymbol{f}_{t}^{b}, & \boldsymbol{\omega}_{ib,t}^{b} \end{bmatrix}^{T} + \boldsymbol{\delta}^{\mathrm{IMU}} \label{eq:imu} \\
    \boldsymbol{Z}_{t}^{\mathrm{GPS}} &= \begin{bmatrix} \phi_{t}^{n}, & \lambda_{t}^{n}, & h_{t}^{n} \end{bmatrix}^{T} + \boldsymbol{\delta}^{\mathrm{GPS}} \label{eq:gps} \\
    \boldsymbol{Z}_{t}^{\mathrm{UWB}} &= \begin{bmatrix} d_{t}^{n}, & \alpha_{t}^{n}, & \beta_{t}^{n} \end{bmatrix}^{T} + \boldsymbol{\delta}^{\mathrm{UWB}} \label{eq:uwb} \\
    \boldsymbol{Z}_{t}^{\mathrm{Baro}} &= h_{t}^{B,n} + \delta^{\mathrm{Baro}} \label{eq:baro} \\
    \boldsymbol{Z}^F_{t} &= \begin{bmatrix} \boldsymbol{Z}_{t}^{\mathrm{IMU}}, & \boldsymbol{Z}_{t}^{\mathrm{GPS}}, & \boldsymbol{Z}_{t}^{\mathrm{UWB}}, & \boldsymbol{Z}_{t}^{\mathrm{Baro}} \end{bmatrix}^{T} \label{eq:total}
\end{align}
\end{subequations}

\noindent where $\boldsymbol{Z}^F_{t}$ represents the combination of observations from each subsystem, and $\boldsymbol{\delta}$ or $\delta$ with a superscript represents observation errors, respectively.

The objective is to minimize the error (to achieve the highest accuracy) and the variance of the error (to achieve the highest robustness) of the final localization result by using processing algorithms and fusion methods, as shown in Equation (2).

\begin{subequations}
\begin{align}
\boldsymbol{X}_{t}^{F} & = \underset{\boldsymbol{Z}_{t}}{\arg\min}\left(k_{1}\left\|\boldsymbol{X}_{t}^{F}-\boldsymbol{X}_{t}^{\text{True}}\right\|_{2}+k_{2}\operatorname{tr}\boldsymbol{\Sigma}_{t}\right) \\
\boldsymbol{X}_{t}^{F} & = f^{F}\left( \boldsymbol{Z}_{t}^{\text{IMU}},\boldsymbol{Z}_{t}^{\text{GPS}},\boldsymbol{Z}_{t}^{\text{UWB}},\boldsymbol{Z}_{t}^{\text{Baro}}\right) \\
\boldsymbol{\Sigma}_{t} & = \mathbb{E}\left[\left( \boldsymbol{X}_{t}^{F}-\bar{\boldsymbol{X}}_{t}\right)\left( \boldsymbol{X}_{t}^{F}-\bar{\boldsymbol{X}}_{t}\right)^{T}\right]
\end{align}
\end{subequations}

\noindent where $f^{F}$ represents the combined algorithmic flow model that integrates the processing algorithms and  fusion methods, and $\boldsymbol{X}_{t}^{F}$ represents the final fusion results.

\section{Fusion architecture model}

To address the localization problem for the high-altitude climbing robot as defined in \textbf{Problem 1} and the scenario model shown in Figure 2, a multi-sensor fusion architecture is designed, as illustrated in Figure 3. In this architecture, GPS provides absolute position information within the navigation coordinate system and is combined with the IMU to form a conventional GPS/INS-EKF model. This partially compensates for the degradation in accuracy and robustness in semi-occluded environments. The planar array UWB, in turn, offers stable localization information in a local coordinate system, enhancing the overall robustness of the localization. The barometer, specializing in altitude measurement [41], effectively remedies the precision deficiencies of GPS in height determination.

\begin{figure}[htbp]
    \centering
    \includegraphics[width=0.48\textwidth]{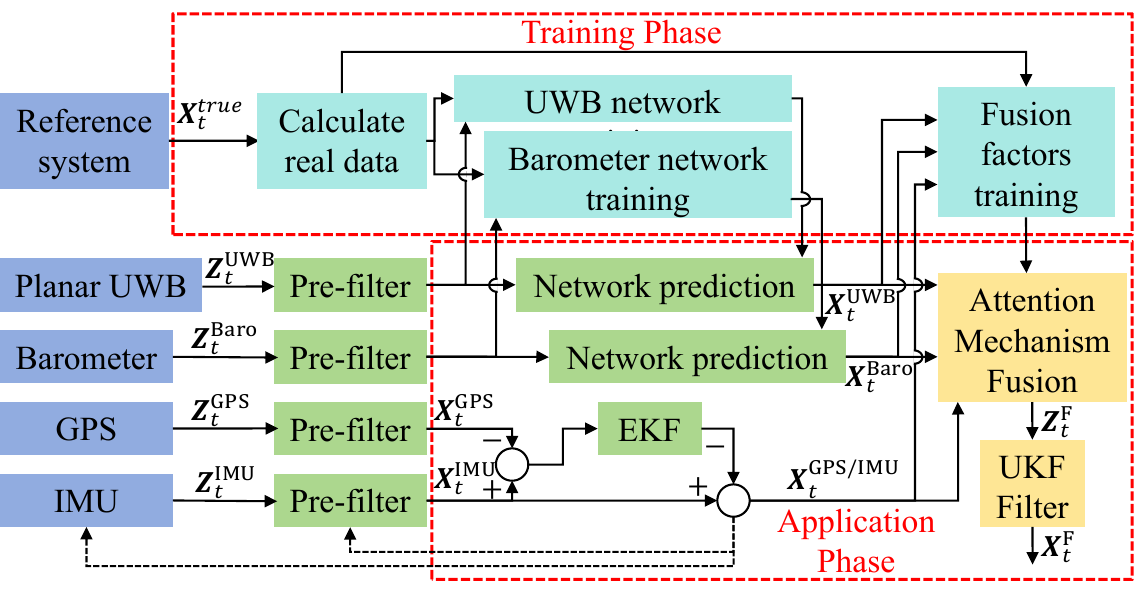}
    \caption{Localization architecture of planar array UWB/GPS/IMU/Barometer Integration}
    \label{f3}
\end{figure}

Furthermore, the integrated architecture in Figure 3 incorporates two end-to-end neural network models and an attention-based fusion mechanism. The models include both training phases and application phases. During the training phase, a large volume of raw measurement data is first collected, including the ground truth data $\boldsymbol{X}^{true}_t$ from a reference system and the raw data $\boldsymbol{Z}_t$ from each subsystem. This data is preprocessed firstly. Then, parameters are trained for three networks: the planar array UWB inference network, the barometer inference network, and the fusion factor inference network.

Once training is complete, the system enters the application phase. Real-time motion data is input into the algorithmic flow processing architecture, and the trained networks are used for prediction and data fusion. Finally, the fused data is processed using an UKF to enhance the accuracy and robustness of the final localization result.

\subsection{Planar Array UWB Localization}

Compared to traditional UWB ranging devices, a planar array UWB integrates multiple UWB antennas on a single anchor, enabling it to perform ranging and direction estimation simultaneously, thus achieving single-anchor localization. Its localization model is shown in Figure 4. The planar array measures the angles with the central axis in two directions, along with the distance, to determine a 3D vector can be converted into position of the target. 

\begin{figure}[htbp]
    \centering
    \includegraphics[width=0.48\textwidth]{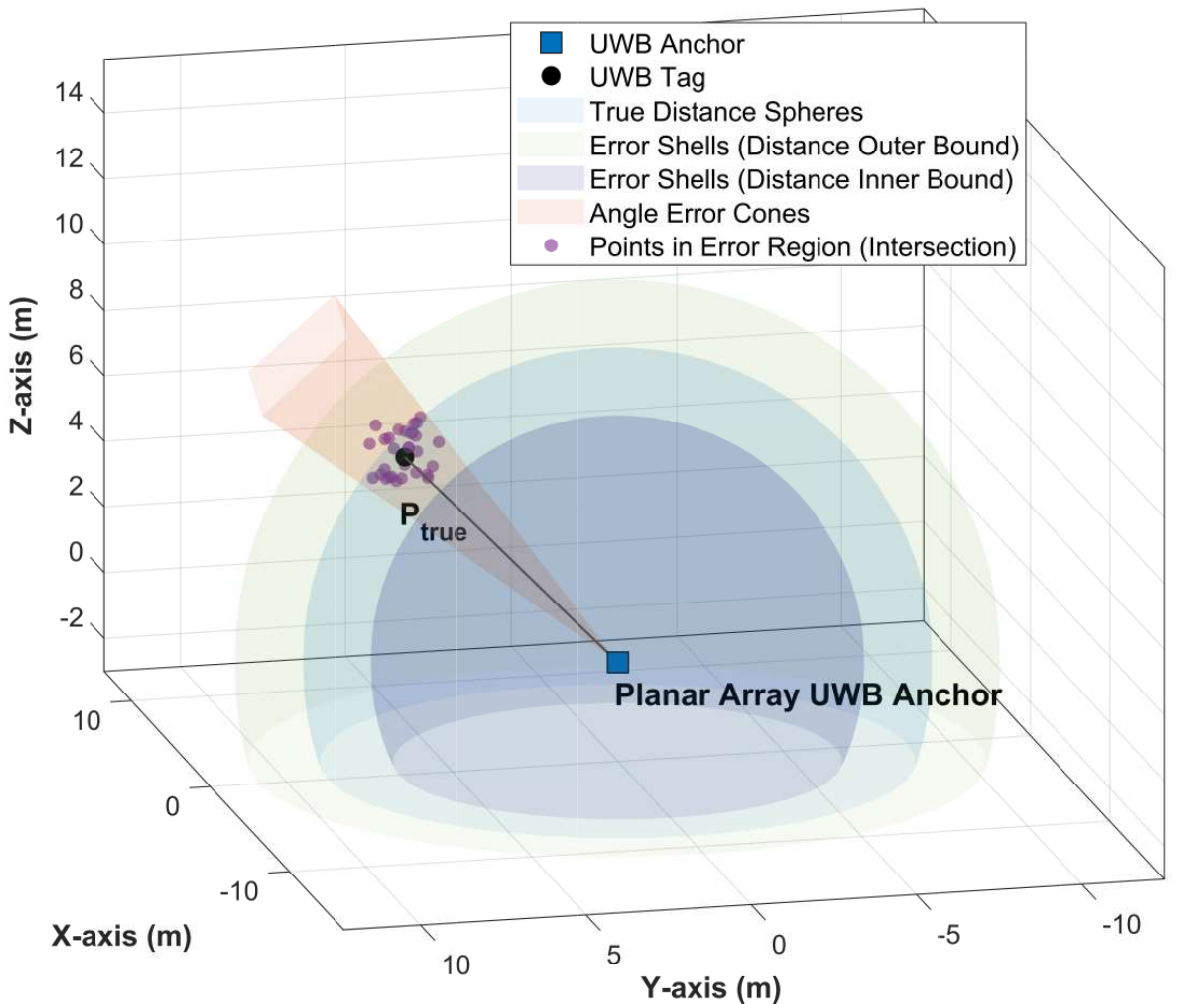}
    \caption{Localization model of Planar Array UWB}
    \label{f4}
\end{figure}

The basic geometric calculation model is shown in Equation (3).

\begin{equation}
\boldsymbol{X}_{t}^{\mathrm{UWB}} = \boldsymbol{\bar{p}_{t}^{n}} +\boldsymbol{R}_{l}^{n}
\left[ \begin{array}{c}
    d_{t}^{l} \sin \alpha_{t}^{l} \\
    d_{t}^{l} \sin \beta_{t}^{l} \\
    d_{t}^{l} \cos \alpha_{t}^{l} \cos \beta_{t}^{l}
\end{array} \right]
\end{equation}

\noindent where $\bar{\boldsymbol{p}}_{t}^{n}$ represents the average position of the antennas on the UWB anchor in navigation frame, $\boldsymbol{R}_{l}^{n}$ denotes the rotation matrix from local frame to  navigation frame.

It can be observed that since most planar array UWB devices can directly output angle and distance information, their calculation model is based on spatial geometry. However, this type of model uses an approximation for angle computation, which results in poor accuracy and high sensitivity to errors. To improve these characteristics, this paper designs an end-to-end calculation model based on a Fully Connected Neural Network (FCNN) that directly learns the position solution error-present conditions. The network model is shown in Figure 5. Because the positioning output of UWB is correlated with measurements from nearby moments in the time series, the model uses a sliding window to choose the measurements from the latest $k$ moments as input. The connection weights of the hidden layer are then trained to output both a position prediction for the current moment and a position error prediction for the traditional model, which preserves the intrinsic reliability characteristics of the data.

\begin{figure}[htbp]
    \centering
    \includegraphics[width=0.45\textwidth]{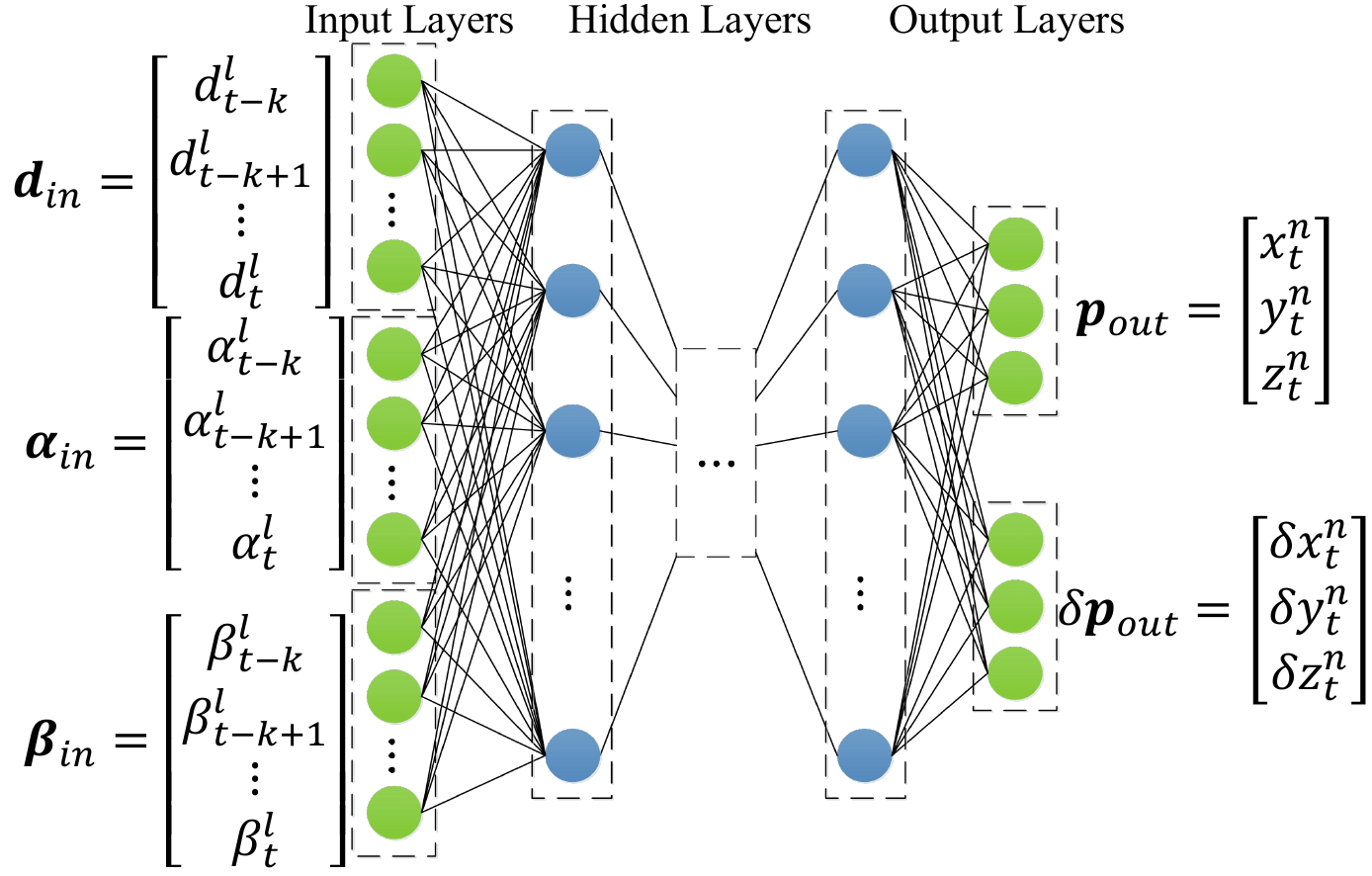}
    \caption{FCNN Prediction Model for Planar Array UWB Solution}
    \label{f5}
\end{figure}

\subsection{GPS/IMU Inertial Navigation Model}

GPS and IMU have complementary characteristics and have been extensively studied and applied. Therefore, this paper directly adopts a mature GPS/INS fusion model to integrate GPS and IMU data, using it as a preliminary input for subsequent fusion. Its state update equation is shown in Equation 4 [42].

\begin{subequations}
\label{eq:gps_imu_error_dynamics}
\begin{align}
\boldsymbol{X}_{t}^{\mathrm{GPS/IMU}} & = 
\left[ \begin{array}{c}
    \boldsymbol{\hat{r}}^{n} \\
    \boldsymbol{\hat{v}}^{n} \\
    \boldsymbol{\hat{C}}_{b}^{n}
\end{array} \right] = 
\left[ \begin{array}{c}
    \boldsymbol{r}^{n} + \delta \boldsymbol{r}^{n} \\
    \boldsymbol{v}^{n} + \delta \boldsymbol{v}^{n} \\
    \left(1 - \boldsymbol{E}^{n}\right) \boldsymbol{C}_{b}^{n}
\end{array} \right] \\
\delta \boldsymbol{\dot{r}}^{n} & = \boldsymbol{F}_{r r} \delta \boldsymbol{r}^{n} + \boldsymbol{F}_{r v} \delta \boldsymbol{v}^{n} \\
\delta \boldsymbol{\dot{v}}^{n} & = -\left(2 \delta \boldsymbol{\omega}_{i e}^{n} + \delta \boldsymbol{\omega}_{e n}^{n}\right) \times \boldsymbol{v}^{n} + \delta g^{n} \nonumber \\
& \quad -\left(2 \boldsymbol{\omega}_{i e}^{n} + \boldsymbol{\omega}_{e n}^{n}\right) \times \delta \boldsymbol{v}^{n}  \nonumber\\
& \quad + \boldsymbol{f}^{n} \times \boldsymbol{e}^{n} + \boldsymbol{C}_{b}^{n} \delta \boldsymbol{f}^{n} \\
\boldsymbol{\dot{\varepsilon}}^{n} & = \boldsymbol{F}_{e r} \delta \boldsymbol{r}^{n} + \boldsymbol{F}_{e v} \delta \boldsymbol{v}^{n} +\nonumber\\
& \quad  \left[\left(\boldsymbol{\omega}_{i e}^{n} + \boldsymbol{\omega}_{e n}^{n}\right) \times\right] \boldsymbol{\varepsilon}^{n} - \boldsymbol{C}_{b}^{n} \delta \boldsymbol{\omega}^{b}
\end{align}
\end{subequations}

Since the GPS/INS-EKF is a mature model, its related definitions are not repeated in this paper\footnote{These symbol definitions can be found in Reference [42].}. 

\subsection{Barometer Altitude Model}

A barometer estimates altitude by measuring atmospheric pressure. It features high sensitivity and high short-term precision but is susceptible to temperature drift and meteorological interference. Its basic altitude calculation model is shown in Equation 5.

\begin{align}
h_{t}^{B,n} = t_{0}/V\left(1-\left(P_{t}/P_{0}\right)^{RV/(gM)}\right)
\end{align}

\noindent where $h_{t}^{B,n}$ represents the altitude calculated by the barometer model, $P_t$ is the current pressure measurement, $P_{0}$ is the reference pressure at a known altitude, $t_{0}$ is the temperature at that reference altitude, $V$ is the vertical temperature gradient, $g$ is the gravitational acceleration, $M$ is the molar mass of air, and $R$ is the universal gas constant.

Similarly, to address the measurement characteristics of the barometer, an end-to-end neural network to perform altitude calculation and calibration directly under error-present conditions was designed. The network model is shown in Figure 6. The input consists of pressure measurements from the previous $k$ time steps, captured using a sliding window, along with the altitude output from the sensor's internal model. The output is the altitude estimate for the current moment and the altitude error estimate of the traditional model (retaining the inherent reliability features of the data).

\begin{figure}[htbp]
    \centering
    \includegraphics[width=0.4\textwidth]{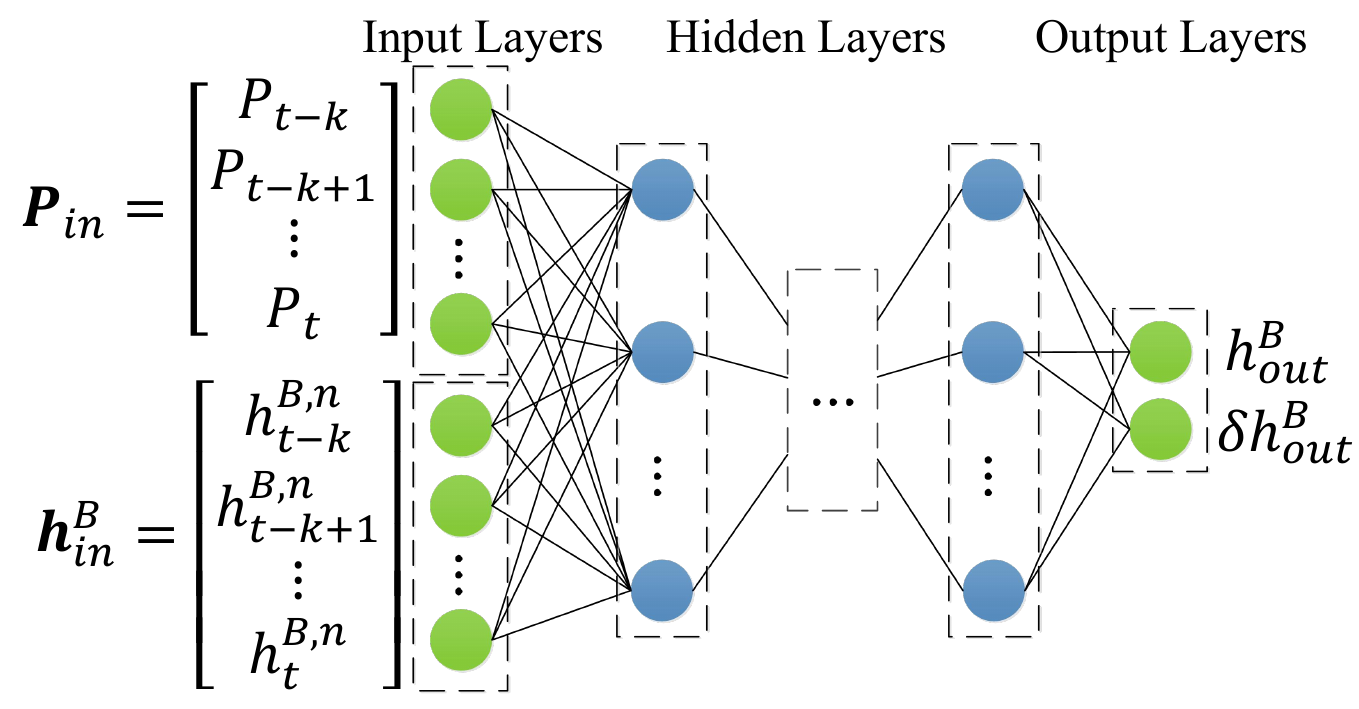}
    \caption{FCNN Prediction Model for Barometer Solution}
    \label{f6}
\end{figure}

\section{multimodal Attention Fusion Mechanism}
Multi-sensor fusion is an established approach to improve localization accuracy and robustness. Classical filtering-based methods such as EKF and Federal Kalman Filter (FKF) can effectively combine heterogeneous measurements, but they typically rely on fixed probabilistic assumptions and may underfit the dynamic, modality-dependent reliability encountered in real environments. To address this limitation, we design a multimodal uncertainty-aware attention fusion model that jointly encodes temporal context, learns fusion ratios via attention with reliability gating, and outputs fused observations with adaptive covariance for UKF updating. The overall architecture is shown in Figure 7.  

\begin{figure}[htbp]
    \centering
    \includegraphics[width=0.49\textwidth]{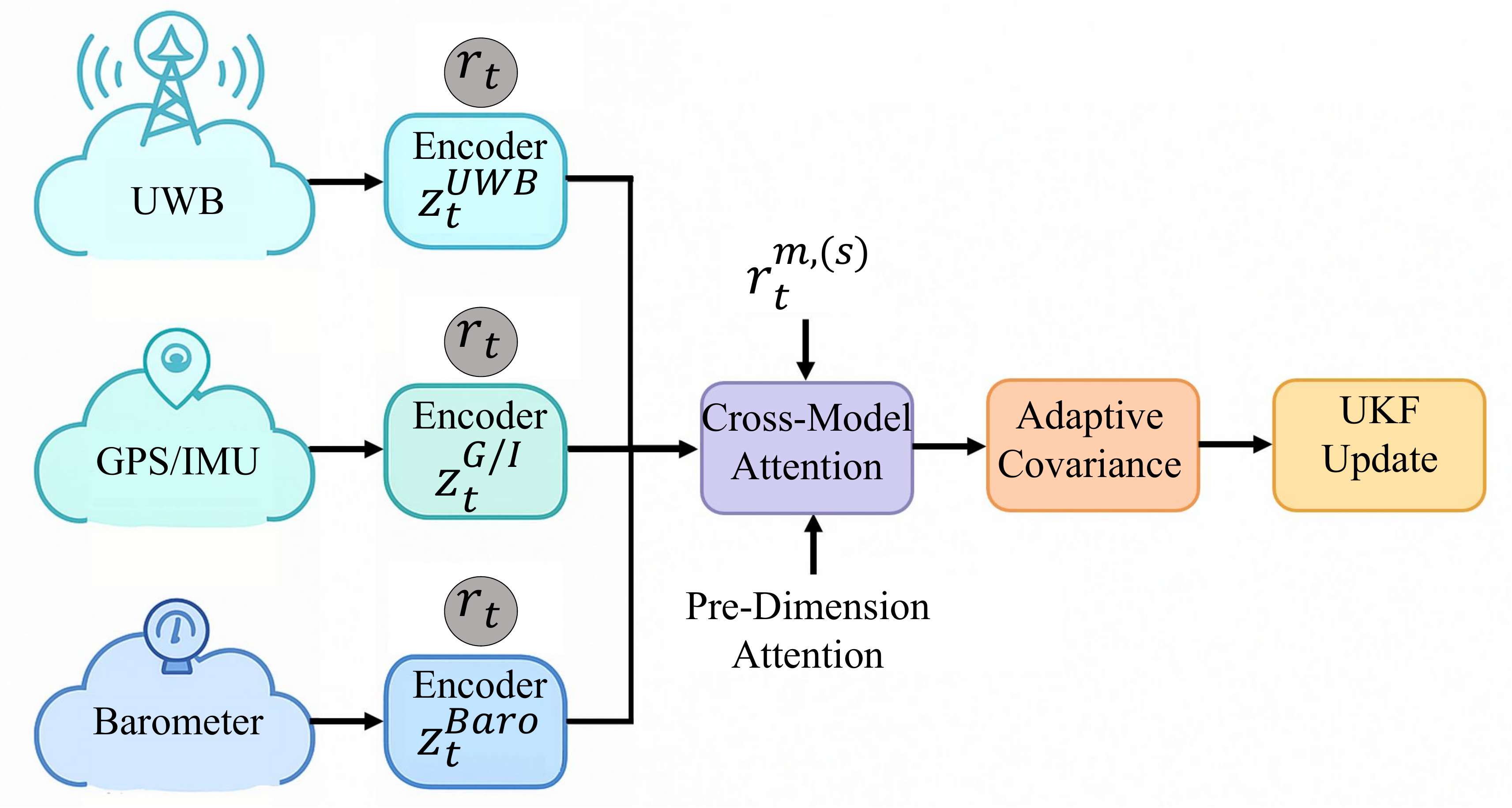}
    \caption{Multimodal Attention Fusion Architecture}
    \label{f7}
\end{figure}

First, given sliding-window observations at time $t$ with length $L$, the inputs are 
$X^{\text{UWB}}_{t-L+1:t}$, 
$X^{\text{GPS/IMU}}_{t-L+1:t}$, and 
$X^{\text{Baro}}_{t-L+1:t}$. 
Each modality is passed through an encoder $E_m(\cdot)$ to obtain a hidden representation:  

\begin{equation}
\mathbf{z}_t^{m}= E_m(X^m_{t-L+1:t}), 
 m \in \{\text{UWB}, \text{GPS/IMU}, \text{Baro}\}
\end{equation}

At the same time, a quality estimation head computes a reliability score for each modality, denoted as $\mathbf{r}_t^m$ (e.g., NLOS confidence for UWB, HDOP for GPS, residual statistics for IMU, and variance for the barometer).  

Next, the cross-modal attention block operates per spatial axis $s\in\{x,y,z\}$. A query is derived from the concatenated embeddings, $\mathbf{q}_t^{(s)} = W_q^{(s)}[\mathbf{z}_t^{\text{UWB}};\mathbf{z}_t^{\text{GPS/IMU}};\mathbf{z}_t^{\text{Baro}}]$, and interacts with keys $\mathbf{k}_t^m=W_k\mathbf{z}_t^m$. The attention logit is defined as  

\begin{equation}
s_t^{m,(s)} = 
\frac{\mathbf{q}_t^{(s)\top}\mathbf{k}_t^m}{\sqrt{d_k}}
+ \beta^{(s)}\mathbf{w}^{(s)\top}\mathbf{r}_t^m
+ b_{\text{prior}}^{m,(s)}
\end{equation}

\noindent where the reliability term and prior bias guide the fusion toward trustworthy modalities (e.g., the barometer on $z$). After normalization, the fusion ratios are obtained as  

\begin{equation}
\gamma_t^{m,(s)} =
\frac{\exp(s_t^{m,(s)})}{\sum_{m'} \exp(s_t^{m',(s)})}
\end{equation}

These ratios serve as convex weights for modality-specific estimates $\hat{x}_t^{m,(s)}$, leading to the fused observation

\begin{subequations}
\begin{align}
\tilde{\mathbf{Z}}^F_t = 
[\tilde{z}_t^{(x)}, \tilde{z}_t^{(y)}, \tilde{z}_t^{(z)}]^\top\\
\tilde{z}_t^{(s)} = \sum_m \gamma_t^{m,(s)} \hat{x}_t^{m,(s)}
\end{align}
\end{subequations}

To enhance robustness, the measurement covariance is adaptively estimated as  

\begin{equation}
\Sigma_t^{(s)} =
\sum_m \gamma_t^{m,(s)} (\sigma_t^{m,(s)})^2
+ \lambda \sum_m \gamma_t^{m,(s)} \big(\hat{x}_t^{m,(s)} - \tilde{z}_t^{(s)}\big)^2
\end{equation}

\noindent where the first term is the weighted intrinsic variance of each modality and the second term penalizes inter-modality divergence when estimates are inconsistent.  

Finally, the fused observation $\tilde{\mathbf{z}}_t$ and the adaptive covariance $\Sigma_t$ are jointly fed into a UKF for state updating, yielding a smooth and reliable trajectory. During training, the model is optimized against RTK ground truth using a weighted regression loss, where the $z$-axis receives higher penalty to emphasize altitude accuracy. Evaluation metrics include axis-wise Mean Absolute Error (MAE) and Root Mean Square Error (RMSE), ensuring unbiased assessment by averaging per-sample Euclidean errors instead of axis-averaged errors.

\section{Experiments}

To evaluate the performance of the models and algorithms described in the paper, a positioning experiment was designed using Liduo Technology’s quadruped climbing robot, which is illustrated in Figure 8. The experimental devices consisted of:
A. Quadruped climbing robot,
B. RTK mobile station, barometer, GPS/IMU integrated sensor, UWB tag, and pressure sensor,
C. Ground control computer system,
D. Array UWB anchor and RTK base station.
The experimental scenario is depicted in E. The safety rope system is fixed to the top of the building, suspending the robot against the external wall. The robot can move using either its foot suction cups or the safety rope mechanism. It is equipped with an RTK mobile station, a GPS/IMU integrated sensor, a UWB tag, and a pressure sensor to collect real-time data during movement. Deployed nearby are a planar array UWB anchor, an RTK base station, and a ground control station.

\begin{figure}[tbp]
    \centering
    \includegraphics[width=0.48\textwidth]{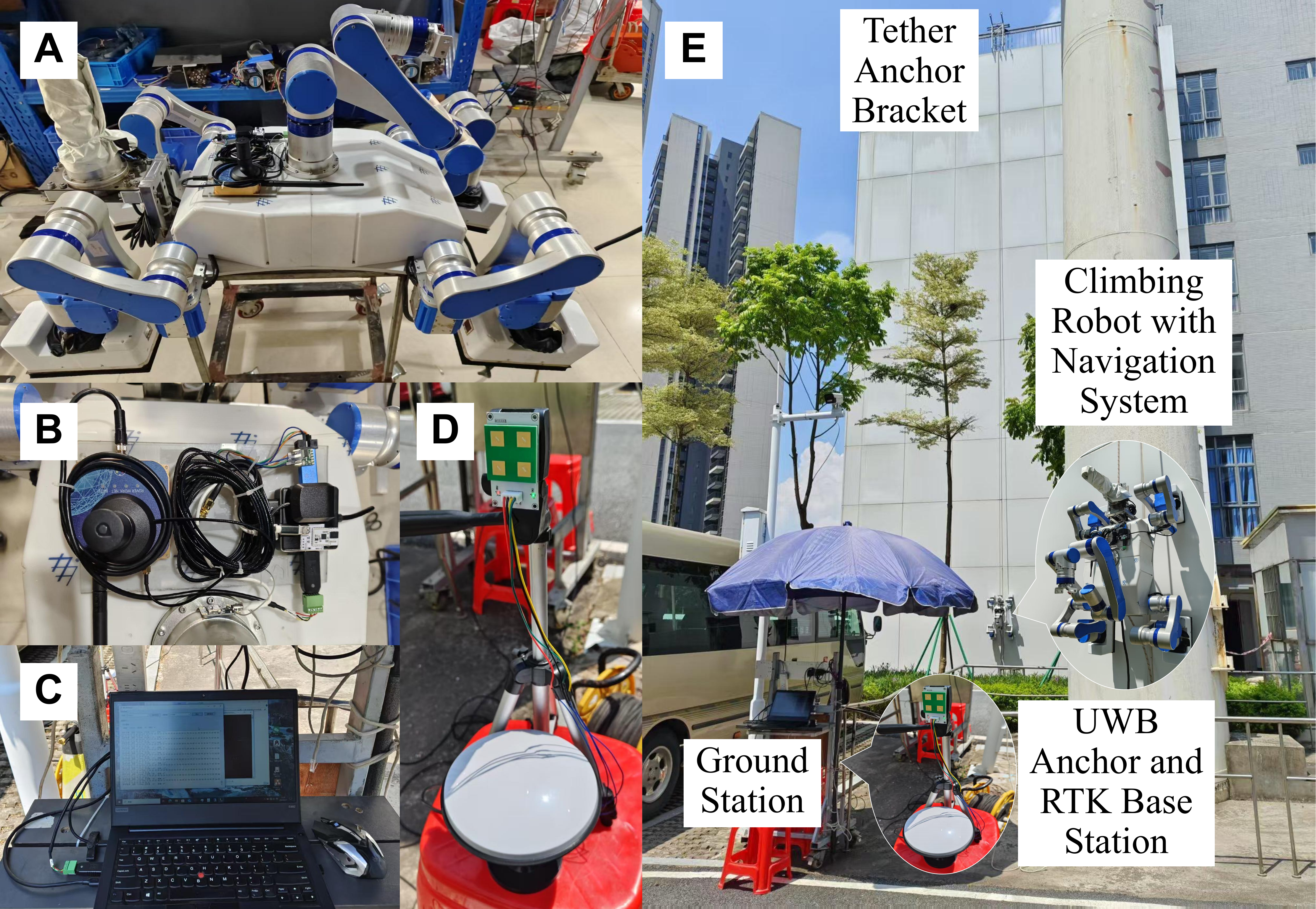}
    \caption{The setup of experimental devices and scenario. A, B, C, D presents devices including climbing robot, integrated UWB/GPS/IMU/Barometer system, RTK mobile station, computer, planar array UWB anchor and RTK base station. E illustrates localization scenario of quadruped climbing robot.}
    \label{f8}
\end{figure}

For the metrics of the algorithms, the paper selects the RMSE, the Standard Deviation (STD) of RMSE (as shown in Equation (2)) and MAX value of RMSE, as performance indicators. Comparative tests were conducted on the following algorithms: (1) Barometer traditional model, (2) Barometer FCNN (ours), (3) UWB geometric algorithm, (4) UWB FCNN (ours), (5) GPS/IMU-EKF, (6) Attention Mechanism-based Fusion Algorithm (AMFA) (ours).

The results of the positioning experiment are shown in Figure 9, including value of the three-axis coordinates under the ENU frame. It can be observed that FCNN-based algorithms (2) and (4) are closer than (1) and (3), but exhibit increased volatility, indicating that FCNNs can replace traditional methods for end-to-end estimation with higher accuracy, though they struggle to capture temporal continuity. Algorithm (5) performs poorly in both accuracy and robustness due to severe GPS occlusion. The proposed method (6) integrates multiple sensors, uses FCNN to enhance accuracy, and employs UKF to mitigate robustness issues, ultimately achieving the smoothest trajectory closest to RTK.

\begin{figure}[tbp]
    \centering
    \includegraphics[width=0.48\textwidth]{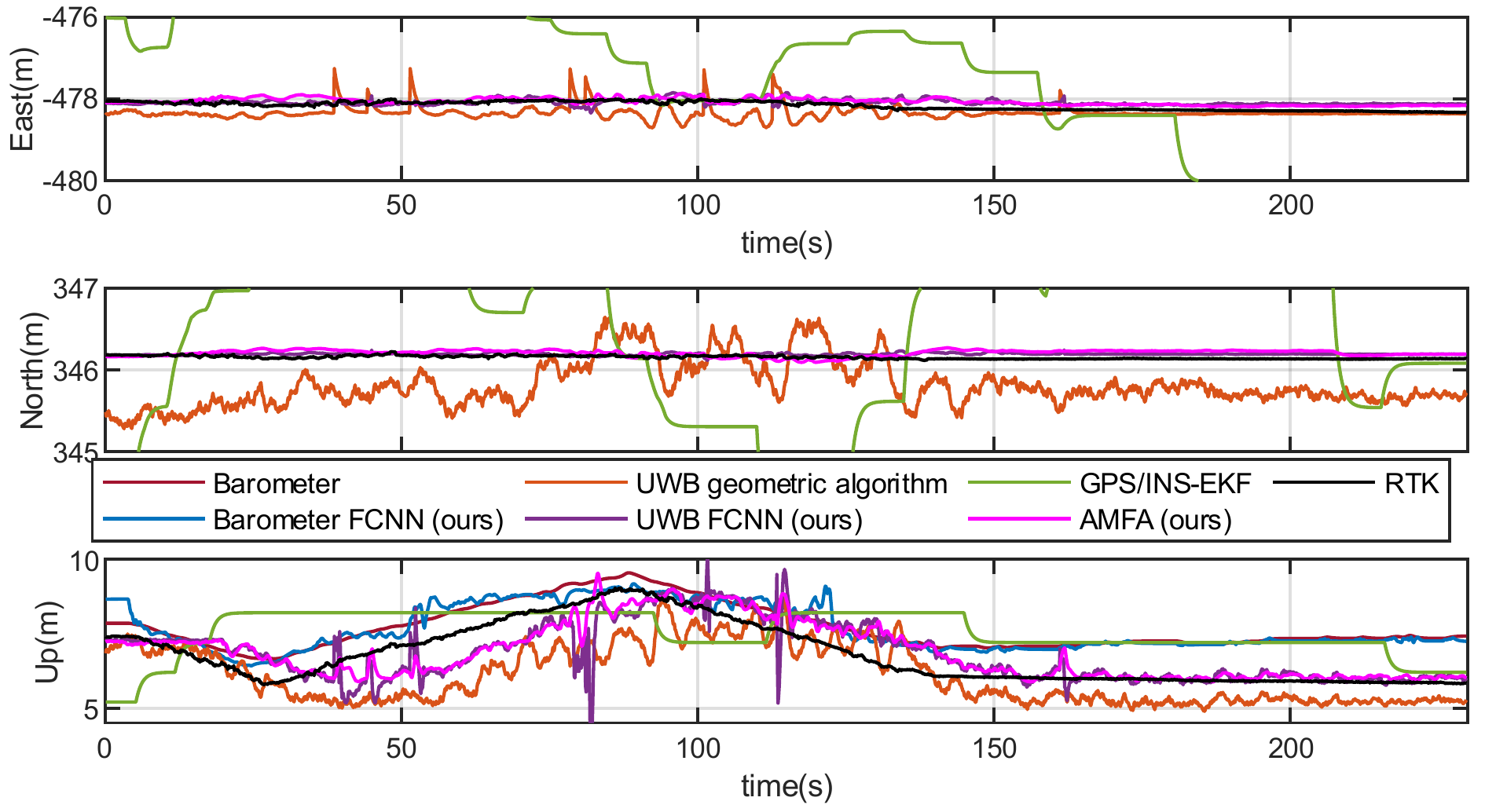}
    \caption{Variations in three-axis coordinate values across different algorithms}
    \label{f9}
\end{figure}

The boxplot comparison of RMSE under the East-North-Up (ENU) frame is presented in Figure 10. The error distributions vary across algorithms, reflecting their inherent characteristics. Specifically, formula-based algorithms (1), (3) and (5) show lower accuracy and high error sensitivity. In contrast, FCNN-based methods (2) and (4) improve accuracy but reduce robustness. The fused architecture of algorithm (6) effectively addresses both accuracy and robustness, achieving optimal performance.

\begin{figure}[htbp]
    \centering
    \includegraphics[width=0.48\textwidth]{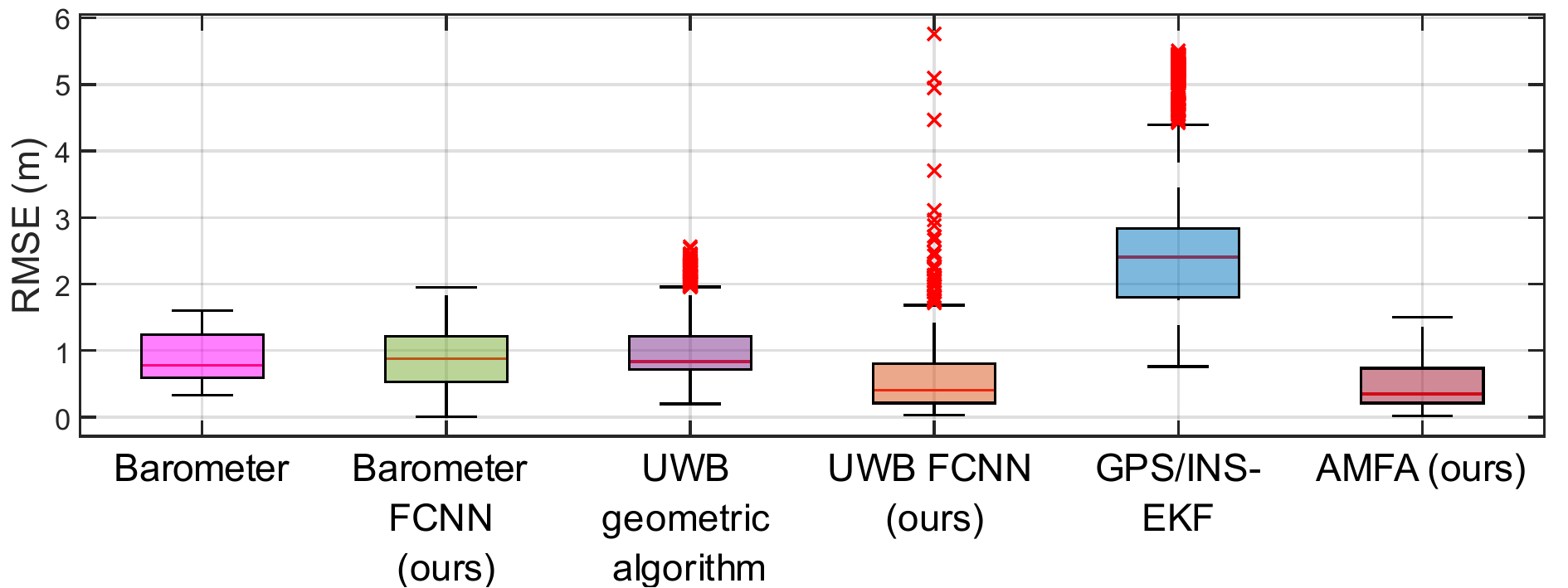}
    \caption{Boxplot comparison of different algorithms}
    \label{f10}
\end{figure}

Table 1 displays a comparison of RMSE, STD of RMSE and Max RMSE between different algorithms. It can be observed that the barometer achieves the optimal STD of 0.31m on the Z-axis. However, in terms of overall three-axis accuracy and stability, the proposed integrated algorithm (6) achieves the best performance across all metrics, with RMSE, STD, and MAX reaching 0.48m, 0.43m, and 1.5m respectively.

\begin{table}[htbp]
\caption{Performance metrics comparison of different algorithms}
\label{table_example}
\begin{center}
\begin{tabular}{|c|c|c|c|}
\hline
Algorithms & RMSE(m) & STD(m) & MAX(m)\\
\hline
(1) Barometer &0.89 &	{0.31}&	1.60\\
\hline
\textbf{(2) Barometer FCNN (ours)} & \textbf{0.85$\downarrow$} &	\textbf{0.35$\uparrow$}& \textbf{1.95$\uparrow$}	\\
\hline
(3) UWB geomitric algorithm  &1.01 &	0.63&2.56	\\
\hline
\textbf{(4) UWB FCNN (ours)} &\textbf{0.55$\downarrow$} &	\textbf{0.52$\downarrow$}&	\textbf{5.76$\uparrow$}\\
\hline
(5) GPS/INS-EKF &2.66 &	2.41& 5.51	\\
\hline
\textbf{(6) AMFA (ours)} & \textbf{0.48$\downarrow$} &	\textbf{0.43$\downarrow$} &	\textbf{1.50$\downarrow$}\\
\hline
\end{tabular}
\end{center}
\end{table}

The Cumulative Distribution Function (CDF) picture of experimental errors is illustrated in Figure 11. The CDF plot reinforces the same conclusions as the boxplot: Algorithm (6) exhibits the steepest curve and converges to 1 the fastest, demonstrating its superior performance. In contrast, Algorithm (5) degrades significantly in the climbing scenario, resulting in the flattest curve. Algorithm (2) performs similarly to Algorithm (1), while Algorithm (4) shows noticeable improvement over Algorithm (3).

\begin{figure}[htbp]
    \centering
    \includegraphics[width=0.48\textwidth]{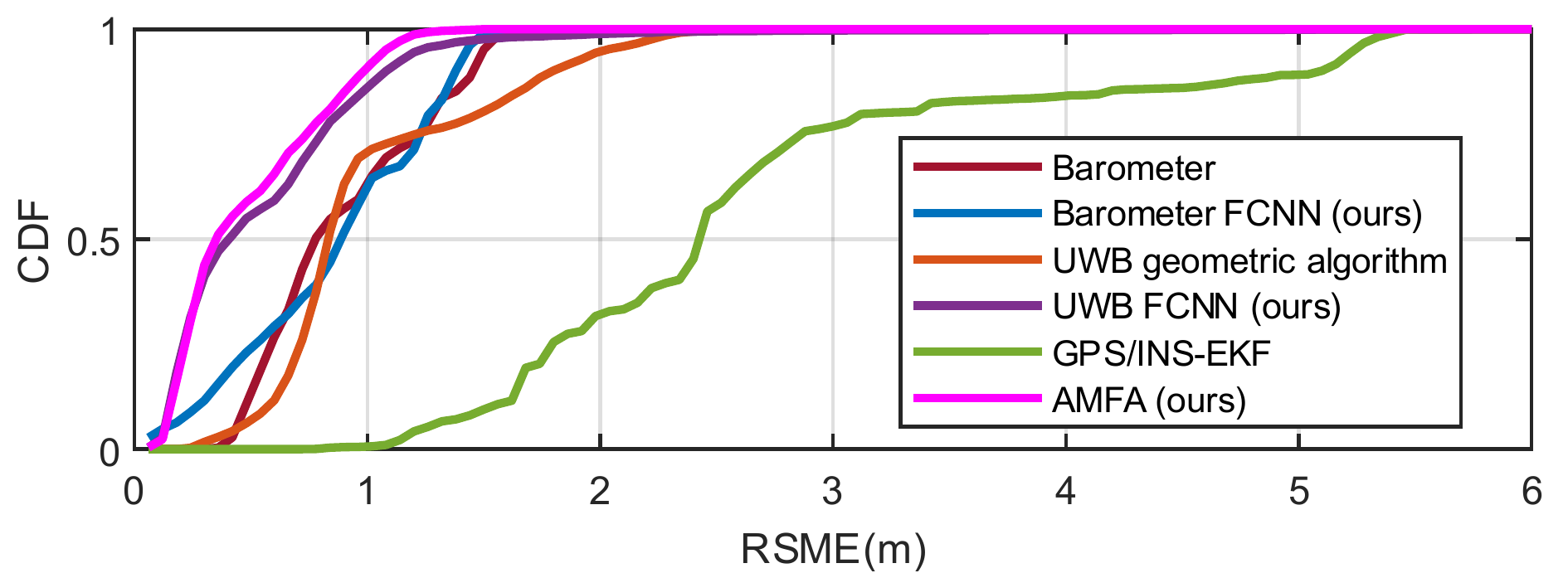}
    \caption{CDF curve comparison of different algorithms}
    \label{f11}
\end{figure}

\section{CONCLUSIONS}

This paper proposes an integrated architecture for high-altitude climbing robot localization, combining multi-sensor data (planar UWB, GPS, IMU, barometer) and algorithms (FCNN, EKF, attention mechanism, UKF) to achieve high-precision continuous localization. Real-world experiments validate its effectiveness, demonstrating its theoretical and practical value for high-altitude robotic navigation.

\addtolength{\textheight}{-1cm}   % This command serves to balance the column lengths

%%%%%%%%%%%%%%%%%%%%%%%%%%%%%%%%%%%%%%%%%%%%%%%%%%%%%%%%%%%%%%%%%%%%%%%%%%%%%%%%

%%%%%%%%%%%%%%%%%%%%%%%%%%%%%%%%%%%%%%%%%%%%%%%%%%%%%%%%%%%%%%%%%%%%%%%%%%%%%%%%



% \section*{ACKNOWLEDGMENT}

% The preferred spelling of the word ÒacknowledgmentÓ in America is without an ÒeÓ after the ÒgÓ. Avoid the stilted expression, ÒOne of us (R. B. G.) thanks . . .Ó  Instead, try ÒR. B. G. thanksÓ. Put sponsor acknowledgments in the unnumbered footnote on the first page.

%%%%%%%%%%%%%%%%%%%%%%%%%%%%%%%%%%%%%%%%%%%%%%%%%%%%%%%%%%%%%%%%%%%%%%%%%%%%%%%%

% References are important to the reader; therefore, each citation must be complete and correct. If at all possible, references should be commonly available publications.


\begin{thebibliography}{99}

\bibitem{c1} K. Tan et al., "A differential geometry modeling method for wheeled climbing robots tracking control on variable curvature surfaces," IEEE Trans. Ind. Electron., vol. 72, no. 7, pp. 7200-7209, Jul. 2025.
\bibitem{c2} Z. Gu, Z. Gong, B. Tao, Z. Yin, and H. Ding, "Global localization based on tether and visual-inertial odometry with adsorption constraints for climbing robots," IEEE Trans. Ind. Informat., vol. 19, no. 5, pp. 6762-6772, May 2023.
\bibitem{c3} M. Magdy, A. Hatem, and N. A. Mostafa, "Analysis of the design parameters of a climbing robot for wind turbine towers inspection," in Proc. 5th Int. Conf. Artif. Intell., Robot. Control (AIRC), Cairo, Egypt, 2024, pp. 102-106.
\bibitem{c4} W. Zhang et al., "Design and development of a new biped robotic system for exoskeleton-structure window cleaning," IEEE Trans. Autom. Sci. Eng., vol. 22, pp. 3160-3171, 2025.
\bibitem{c5} W. Zhang, Z. Li, L. M. Tam, and Q. Xu, "Novel complete coverage path planning of a biped wall-climbing robot for window cleaning," in Proc. IEEE Int. Conf. Robot. Biomimetics (ROBIO), Bangkok, Thailand, 2024, pp. 669-674.
\bibitem{c6} S. Hong et al., "Agile and versatile climbing on ferromagnetic surfaces with a quadrupedal robot," Sci. Robot., vol. 7, p. eadd1017, 2022.
\bibitem{c7} K. Huang et al., "Researches on a wall-climbing robot based on electromagnetic adsorption," in Proc. IEEE 3rd Inf. Technol., Netw., Electron. Autom. Control Conf. (ITNEC), Chengdu, China, 2019, pp. 644-647.
\bibitem{c8} A. Rosyid and B. El-Khasawneh, "A large-scale suction-based climbing parallel robot for wall painting application," in Proc. IEEE Int. Conf. Robot. Autom. (ICRA), Yokohama, Japan, 2024, pp. 2119-2125.
\bibitem{c9} A. Matsubasa, T. Asanuma, and H. Kimura, "Steel wall climbing robot using pneumatic cylinder and internally balanced magnetic unit," in Proc. SICE Festival Annu. Conf. (SICE FES), Kochi City, Japan, 2024, pp. 901-908.
\bibitem{c10} H. Zhu, J. Lu, S. Gu, S. Wei, and Y. Guan, "Planning three-dimensional collision-free optimized climbing path for biped wall-climbing robots," IEEE/ASME Trans. Mechatronics, vol. 26, no. 5, pp. 2712-2723, Oct. 2021.
\bibitem{c11} S. Zhang et al., "Design of magnetic feet for climbing robot with novel circular Halbach net EPM structure," in Proc. 50th Annu. Conf. IEEE Ind. Electron. Soc. (IECON), Chicago, IL, USA, 2024, pp. 1-6.
\bibitem{c12} A. Gawel et al., "A fully-integrated sensing and control system for high-accuracy mobile robotic building construction," in Proc. IEEE/RSJ Int. Conf. Intell. Robots Syst. (IROS), Macau, China, 2019, pp. 2300-2307.
\bibitem{c13} J. Nubert, S. Khattak, and M. Hutter, "Graph-based multi-sensor fusion for consistent localization of autonomous construction robots," in Proc. IEEE Int. Conf. Robot. Autom. (ICRA), Philadelphia, PA, USA, 2022, pp. 10048-10054.
\bibitem{c14} M. J. Mehmood et al., "Design and implementation of a localization app to achieve sub-meter level accuracy using Suparco's Pak-Rehber precise localization service," in Proc. 26th Int. Multi-Topic Conf. (INMIC), Karachi, Pakistan, 2024, pp. 1-6.
\bibitem{c15} H. Huang and W. Liu, "Design of localization accuracy improvement system for robot navigation based on dead reckoning collaboration," in Proc. 4th Int. Conf. Electron., Integr. Circuits Commun. Technol. (EICCT), Chengdu, China, 2025, pp. 490-493.
\bibitem{c16} C. Fan, S. Wei, and Y. Yang, "Application of multi-sensor fusion precise localization and autonomous navigation technology in substation intelligent inspection robot," in Proc. Int. Conf. Elect. Drives, Power Electron. Eng. (EDPEE), Athens, Greece, 2024, pp. 624-629.
\bibitem{c17} R. Aparna et al., "IMU based tracking of a person using nonlinear autoregressive exogenous (NARX) algorithm in GPS-denied areas," in Proc. 1st IEEE Int. Conf. Meas., Instrum., Control Autom. (ICMICA), Kurukshetra, India, 2020, pp. 1-4.
\bibitem{c18} S. Matsuzaki and Y. Hasegawa, "Learning crowd-aware robot navigation from challenging environments via distributed deep reinforcement learning," in Proc. IEEE Int. Conf. Robot. Autom. (ICRA), Philadelphia, PA, USA, 2022, pp. 4730-4736.
\bibitem{c19} J. T. Jang, A. Santamaria-Navarro, B. T. Lopez, and A.-a. Agha-mohammadi, "Analysis of state estimation drift on a MAV using PX4 autopilot and MEMS IMU during dead-reckoning," in Proc. IEEE Aerosp. Conf., Big Sky, MT, USA, 2020, pp. 1-11.
\bibitem{c20} T. Qin, P. Li, and S. Shen, "VINS-Mono: A robust and versatile monocular visual-inertial state estimator," IEEE Trans. Robot., vol. 34, no. 4, pp. 1004-1020, Aug. 2018.
\bibitem{c21} Y. Yao, Y. Liu, Z. Zhou, and X. Xu, "A magnetic interference detection-based fusion heading estimation method for pedestrian dead reckoning localization," IEEE Sensors J., vol. 23, no. 1, pp. 677-688, Jan. 2023.
\bibitem{c22} H. Zhang and C. Ye, "Sampson distance: A new approach to improving visual-inertial odometry's accuracy," in Proc. IEEE/RSJ Int. Conf. Intell. Robots Syst. (IROS), Prague, Czech Republic, 2021, pp. 9184-9189.
\bibitem{c23} J. Li et al., "A low-cost, portable mobile mapping system integrating lidar, smartphone, and GNSS-RTK," arXiv preprint arXiv:2506.15983, 2025.
\bibitem{c24} J. Guo et al., "Autonomous flight control design based on multi-sensor fusion for a low-cost quadrotor in GPS-denied environments," in Proc. 7th Asia-Pac. Conf. Intell. Robot Syst. (ACIRS), Tianjin, China, 2022, pp. 53-57.
\bibitem{c25} Z. Ni et al., "UWB non-line-of-sight recognition and suppression methodology," in Proc. 15th Int. Conf. Commun. Softw. Netw. (ICCSN), Shenyang, China, 2023, pp. 313-318.
\bibitem{c26} P. Agarwal et al., "Comparative analysis of machine learning algorithms for LOS/NLOS identification," in Proc. 1st Int. Conf. Electron., Commun. Signal Process. (ICECSP), New Delhi, India, 2024, pp. 1-5.
\bibitem{c27} A. F. Majeed et al., "Multiclass identification of NLOS conditions in UWB localization based machine learning methods," in Proc. IEEE Int. Conf. Adv. Telecommun. Netw. Technol. (ATNT), Johor Bahru, Malaysia, 2024, pp. 1-4.
\bibitem{c28} S. Zhang, E. Wang, Z. Zhu, J. Yi, Y. Wang and E. Kuai, "UKF-FNN-RIC: A highly accurate UWB localization algorithm for TOA scenario," IEEE Trans. Instrum. Meas., vol. 73, pp. 1-13, 2024, Art no. 8508013, doi: 10.1109/TIM.2024.3476531.
\bibitem{c29} E. Wang et al., "High-precision UWB TDOA localization algorithm based on UKF-FNN-CHAN-RIC," IEEE Trans. Instrum. Meas., vol. 74, pp. 1-13, 2025, Art no. 8506813, doi: 10.1109/TIM.2025.3554908.
\bibitem{c30} G.-J. Gordebeke et al., "Time-domain-optimized antenna array for high-precision IR-UWB localization in harsh urban shipping environments," IEEE Sensors J., vol. 24, no. 5, pp. 5561-5577, Mar. 2024.
\bibitem{c31} W. Zeng, J. Zhang, and T. Zhang, "A robust machine learning based UWB AOA estimation method," in Proc. IEEE 99th Veh. Technol. Conf. (VTC2024-Spring), Singapore, Singapore, 2024, pp. 1-5.
\bibitem{c32} B. Van Herbruggen et al., "Single anchor localization by combining UWB angle-of-arrival and two-way-ranging: an experimental evaluation of the DW3000," in Proc. Int. Conf. Localization GNSS (ICL-GNSS), Antwerp, Belgium, 2024, pp. 1-7.
\bibitem{c33} D. Märzinger et al., "Time-multiplexed AoA estimation and ranging," in Proc. Int. Conf. Localization GNSS (ICL-GNSS), Castellón, Spain, 2023, pp. 1-7.
\bibitem{c34} N. I. Petukhov et al., "Synthesis and performance analysis of UKF for processing measurements of UWB ToF/AoA LNS with long baseline and IMU," in Proc. IEEE XVI Int. Sci. Tech. Conf. Actual Problems Electron. Instrum. Eng. (APEIE), Novosibirsk, Russian Federation, 2023, pp. 780-785.
\bibitem{c35} S. Zhang, H. Tang, L. Chen and Y. Gao, "A seamless pedestrian localization system based on GNSS/IMU/UWB/map integration," IEEE Internet Things J., vol. 12, no. 11, pp. 16653-16667, Jun. 2025, doi: 10.1109/JIOT.2025.3534190.
\bibitem{c36} K. Koide et al., "Tightly coupled range inertial localization on a 3D prior map based on sliding window factor graph optimization," in Proc. IEEE Int. Conf. Robot. Autom. (ICRA), Yokohama, Japan, 2024, pp. 1745-1751.
\bibitem{c37} P. Lyu, B. Wang, J. Lai, S. Bai, M. Liu, and W. Yu, "A factor graph optimization method for high-precision IMU-based navigation system," IEEE Trans. Instrum. Meas., vol. 72, pp. 1-12, 2023.
\bibitem{c38} Cohen, N., and I. Klein, "Adaptive Kalman-Informed Transformer," Eng. Appl. Artif. Intell., vol. 146, 2025.
\bibitem{c39} Y. Kang, J. Ji, H. Xu, Y. Yang, P. Chen, and H. Zhao, "Swin-CDSA: The semantic segmentation of remote sensing images based on cascaded depthwise convolution and spatial attention mechanism," IEEE Geosci. Remote Sens. Lett., vol. 21, pp. 1-5, 2024.
\bibitem{c40} J. T. Jang, A. Santamaria-Navarro, B. T. Lopez, and A.-a. Agha-mohammadi, "Analysis of state estimation drift on a MAV using PX4 autopilot and MEMS IMU during dead-reckoning," in Proc. IEEE Aerosp. Conf., Big Sky, MT, USA, 2020, pp. 1-11.
\bibitem{c41} S. Zhang et al., "A low-cost UAV swarm relative positioning architecture based on BDS/barometer/UWB," IEEE Sens. J., vol. 24, no. 23, pp. 39659-39668, Dec. 2024, doi: 10.1109/JSEN.2024.3476286.
\bibitem{c42} N.-P. Binh and V.-H. Nguyen, "Loosely coupled GPS/INS integration for road vehicles," in Proc. 7th Nat. Sci. Conf. Appl. New Technol. Green Build. (ATiGB), Da Nang, Vietnam, 2022, pp. 43-50.
\end{thebibliography}
\end{document}